\documentclass[10pt,twocolumn,letterpaper]{article}

\usepackage{cvpr}              % To produce the CAMERA-READY version
\usepackage[dvipsnames]{xcolor}

\usepackage{amsmath}
\usepackage{amssymb}
\usepackage{enumitem}
\usepackage{multirow}
\usepackage{booktabs} % for professional tables
\usepackage{makecell}
\usepackage[labelformat=simple]{subcaption}

\definecolor{cvprblue}{rgb}{0.21,0.49,0.74}
\usepackage[pagebackref,breaklinks,colorlinks,citecolor=cvprblue]{hyperref}

\def\paperID{373} % *** Enter the Paper ID here
\def\confName{3DV\xspace}
\def\confYear{2024\xspace}

\title{A Cross Branch Fusion-Based Contrastive Learning Framework for \\ Point Cloud Self-supervised Learning}

\author{Chengzhi Wu$^{1}$ \quad Qianliang Huang$^{1}$ \quad Kun Jin$^{2}$ \quad Julius Pfrommer$^{3}$ \quad Jürgen Beyerer$^{1,3}$ 
\vspace{1pt}
\and
$^{1}$Institute for Anthropomatics and Robotics, Karlsruhe Institute of Technology, Germany \\
$^{2}$Department of Automation, Tsinghua University, China \\
$^{3}$Fraunhofer Institute of Optronics, System Technologies and Image Exploitation IOSB, Germany 
\and
\tt\footnotesize
chengzhi.wu@kit.edu, \quad qianliang.huang@student.kit.edu, \quad 
kun6jin@gmail.com, \\
\tt\footnotesize
\{julius.pfrommer, juergen.beyerer\}@iosb.fraunhofer.de
}

\begin{document}
\maketitle

\begin{abstract}
Contrastive learning is an essential method in self-supervised learning. It primarily employs a multi-branch strategy to compare latent representations obtained from different branches and train the encoder. In the case of multi-modal input, diverse modalities of the same object are fed into distinct branches. When using single-modal data, the same input undergoes various augmentations before being fed into different branches. However, all existing contrastive learning frameworks have so far only performed contrastive operations on the learned features at the final loss end, with no information exchange between different branches prior to this stage. In this paper, for point cloud unsupervised learning without the use of extra training data, we propose a \textbf{C}ontrastive \textbf{C}ross-branch \textbf{A}ttention-based framework for \textbf{Po}int cloud data (termed PoCCA), to learn rich 3D point cloud representations. By introducing sub-branches, PoCCA allows information exchange between different branches before the loss end. Experimental results demonstrate that in the case of using no extra training data, the representations learned with our self-supervised model achieve state-of-the-art performances when used for downstream tasks on point clouds.

\end{abstract}    
\section{Introduction}
\label{sec:intro}
Contrastive learning stands as a pivotal method for learning latent representations, especially in the domain of computer vision and natural language processing. However, while its success has been widely demonstrated in 2D image-based tasks \cite{He2020MomentumCF, Chen2020ASF, richemond2020byol, chen2021exploring}, its application to self-supervised contrastive learning on 3D point cloud data has remained relatively underexplored.
Point clouds possess unique characteristics and structural complexities that necessitate tailored approaches for representation learning. Despite the challenges, and while the majority still use reconstruction-based methods \cite{Radford2016UnsupervisedRL, Chen2016InfoGANIR, Brock2019LargeSG, Karras2018ProgressiveGO, Bao2022BEiTBP, He2022MaskedAA, Xie2022SimMIMAS}, recent research has started to address this gap. Promising advancements have been made in the domain of contrastive learning-based point cloud self-supervised learning. 
For example, STRL \cite{Huang2021SpatiotemporalSR} extends BYOL \cite{richemond2020byol} to the 3D domain and performs contrastive learning on global representations directly. PointContrast \cite{Xie2020PointContrastUP} performs contrastive learning on the point-wise level yet is computationally expensive. While Info3D \cite{info3d} uses the shape part and the full shape as positive pairs directly, HSN \cite{hns} only considers the part pair information and ignores the information of the whole shape. 

On the other hand, in recent years, the surge of large models has sparked significant interest in multi-modality-based approaches that leverage extra training data for contrastive-based point cloud representation learning. For example, CrossPoint \cite{Afham2022CrossPointSC} and I2P-MAE \cite{I2P-MAE} use multi-view images of the objects as extra training data, while ReCon \cite{ReCon} further introduces additional text information for richer latent representation learning. 
However, there are still many cases that only single modality data is available for the training. 
Moreover, in all the above methods, latent representations are learned solely on each branch, with no information exchange before the loss end. 
In this paper, we rethink the way of applying contrastive learning to point clouds without extra training data, and explore the new possibility of incorporating information from different branches when using a multi-branch framework.

The key idea of contrastive learning is to use multiple branches to learn multiple latent representations for the same input of different variants, and the network is trained by minimizing their latent representation differences.
When multi-modality is used, it is natural to use one branch for each modality. For example, CrossPoint \cite{Afham2022CrossPointSC} uses one branch for images and the other branch for point clouds. The encoders for each modality on the perspective branches are totally different and the whole framework can be trained stably. 
When only one modality is used, the most widely used method is to perform different augmentations on the same input and use augmented variants for different branches \cite{Huang2021SpatiotemporalSR}.
However, the model may collapse easily if the vanilla model of both branches sharing a same encoder is used, i.e., the encoder encodes everything into a same latent representation. Various methods have been proposed to deal with this problem, including introducing negative pairs \cite{Chen2020ASF, He2020MomentumCF}, using memory bank \cite{InstDisc, Wei2022LearningGP}, adding a predictor on single branch \cite{richemond2020byol, chen2021exploring}, and using momentum update for certain branch encoder \cite{He2020MomentumCF, richemond2020byol, Huang2021SpatiotemporalSR}, etc.
In our case, we only use a single modality of point cloud data and only use positive pairs for training. Following BYOL \cite{richemond2020byol}, the single-branch predictor strategy and the momentum update strategy are adopted to prevent model collapse.

\begin{figure*}[t]
    \centering
    \includegraphics[width=\linewidth,trim=2 2 2 2,clip]{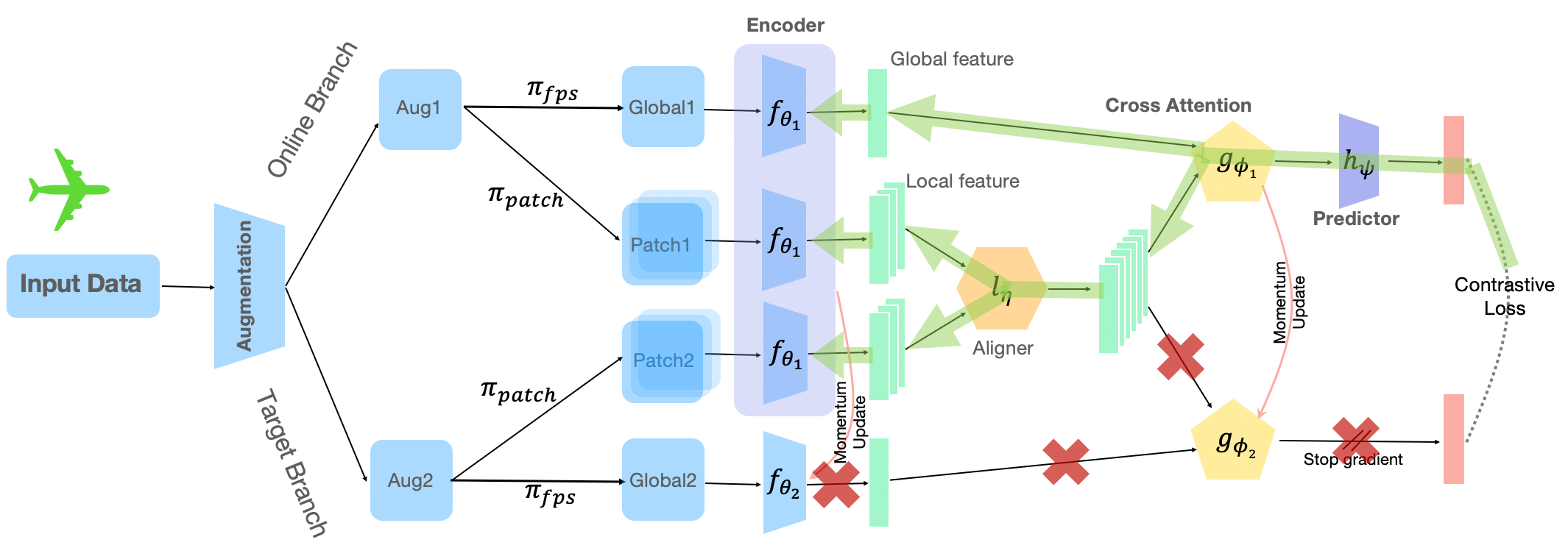}
    \caption{The framework of our proposed PoCCA. Given an input point cloud, it is first augmented with two different augmentation operations. After global sampling and local patching, global and local features are obtained respectively with a pre-designed encoder (e.g. PointNet, DGCNN). The local features from the two branches are then aligned and sent to the cross-attention module to enhance the online- and target global features, respectively. Finally, by comparing the difference between the output representations, we obtain a contrastive loss and train the whole model. See more module details in subsection \ref{sec:methods}.}
    \label{fig:framework}
\end{figure*}

By using each branch to learn one latent representation solely, all existing contrastive learning frameworks only performed contrastive operations on the learned features at the final loss end, with no information exchange between different branches prior to this stage.
At this point, a question arises: is it possible to incorporate information from different branches before the loss end? In this paper, we explore this possibility by introducing sub-branches on the common frameworks. 
To this end, we propose a self-supervised contrastive learning framework for point clouds by fusing (i) the online branch information and the target branch information; (ii) the global sub-branch information and the local sub-branch information, as illustrated in Figure \ref{fig:framework}. PoCCA is a symmetric 3D point cloud representation learning framework. 
It augments the raw point clouds and then samples them globally or locally on different sub-branches. The features from different branches are fused subsequently.
Using different augmentations of the same raw point cloud as a positive pair, the contrastive loss is defined by their latent representation difference. 
Further details of the framework are given in Section \ref{sec:methodology}.
Experimental results demonstrate that the representations learned with our self-supervised model achieve excellent performances when used for downstream tasks on point clouds.

We summarize our contributions as follows:
\begin{itemize}[itemsep=0pt,topsep=-3pt,left=3pt]
 \item A contrastive learning framework that enables information exchange between the online branch and the target branch by introducing sub-branches.
 \item Both global and local features of the input point cloud are extracted. Cross-attention modules are used for global-local feature fusion.
 \item Multiple variants of the proposed contrastive learning framework are evaluated in the ablation study, as well as various local patch sampling methods.
 \item Excellent experimental results on multiple downstream tasks. Among the point cloud unsupervised learning methods that do not use extra training data, PoCCA achieves state-of-the-art results.
\end{itemize}
\section{Related Work}
\label{sec:relatedWork}

%-------------------------------------------------------------------------
\subsection{Contrastive Learning}
Self-supervised learning methods have made great strides in recent years. They are usually either generative-based  \cite{Chu2017CycleGANAM, Karras2019ASG} or contrastive-based  \cite{swav, He2020MomentumCF, richemond2020byol, chen2021exploring}.
Contrastive learning, in contrast to generative models, is a discriminative strategy that tries to separate varied samples  while grouping similar samples. As a pioneering work, InstDisc \cite{InstDisc} separates the extracted features in high dimensional space and categorizes the features as positive and negative samples.
Contrastive Multiview Coding \cite{CMC} extends the definition of positive samples by considering different views from the same object as positive pairs.
MoCo \cite{He2020MomentumCF} proposes a strategy of stopping-gradient on the target branch and instead uses a moving-averaged encoder.
SwAV \cite{swav} further discards the negative samples and compares new sample features with the center of positive features clustering. SimCLR \cite{Chen2020ASF} demonstrates that unsupervised contrastive learning benefits from stronger data augmentations, and a non-linear transformation between the representation and the contrastive loss can significantly improve the quality of the learned representations. SimSam \cite{chen2021exploring} and BYOL \cite{richemond2020byol} further add a predictor at the end and both use only positive pairs for self-supervised learning.

\subsection{Self-supervised Learning on Point Cloud}
In order to accomplish self-supervised representation learning on 3D point clouds, various methods have been proposed. 
Pretext tasks-based self-supervised learning methods are first explored. Jigsaw \cite{jigsaw} is trained by reconstructing point clouds from randomly rearranged parts. PointRotation \cite{rotation} learns point cloud representations by predicting their rotation. STRL \cite{Huang2021SpatiotemporalSR} takes two temporally-correlated frames from a 3D point cloud sequence as the input, transforms it with the spatial data augmentation, and learns the invariant representation in a self-supervised manner. Contrastive-based learning has also been used in several works.
Wang et al.  \cite{occo} pre-trains an encoder with occluded points for downstream tasks. 
Point-level invariant mapping is carried out by PointContrast \cite{Xie2020PointContrastUP} on two transformed views of the input point cloud. 
CrossPoint \cite{Afham2022CrossPointSC} performs cross-modality contrastive learning between point clouds and their corresponding rendered images.
More recently, MAE-based reconstruction methods have shown promising results on point cloud self-supervised learning. Point-BERT \cite{PointBERTP} and Point-MAE \cite{PointMAE} are the first two methods that transfer the idea of MAE to point clouds by masking point patches. A similar encoder is used in MaskPoint \cite{MaskPoint}, but instead of using the transformer as a decoder for reconstruction, it uses the transformer as a discriminator. Point-M2AE \cite{PointM2AE} further updates the encoder with a hierarchical Transformer by introducing multi-scale masking. Using extra training data, I2P-MAE \cite{I2P-MAE} and ReCon \cite{ReCon} obtain superior 3D representations via cross-modal training.

%-------------------------------------------------------------------------
\subsection{Attention Mechanism on Point Cloud}

In recent years, attention-based methods start dominating the image learning domain since ViT  \cite{Dosovitskiy2021AnII}.
More recently, the attention mechanism  \cite{Vaswani2017AttentionIA} has also been proven to be effective for point cloud learning. 
PCT \cite{Guo2021PCTPC} employs self-attention for point cloud understanding with proposed offset-attention.
By constructing a residual point transformer block with self-attention-based layers and linear projections, PT \cite{Zhao2021PointT} builds a U-Net-like network structure. Pointformer \cite{pointformer} proposes a local-global Transformer to integrate features from different levels. 3DPCT \cite{3dpct} designs a dual transformer approach and builds a hierarchical encoder-decoder network. SA-Det3D \cite{sadat} proposes a generic globally-adaptive context aggregation module and a scalable self-attention variant is designed. Pyramid Point Cloud Transformer  \cite{pyratransformer} develops a pyramid module to aggregate multi-scale features. SeedFormer \cite{seedformer} and PoinTr \cite{pointtr} employ attention-based methods for point cloud completion.
APES \cite{Wu2023APES} uses the attention map for sampling edge points of the point clouds.
PatchFormer \cite{patchformer} proposes patch-attention and a lightweight multi-scale attention block. Stratified Transformer  \cite{straftrans} is proposed to additionally sample distant points as keys to capture long-range contexts and demonstrates strong generalization ability. 

\section{Methodology}
\label{sec:methodology}

\subsection{Preliminaries}
\label{sec:pre}
Denote two main branches as branch $A$ and $B$ (in most papers, they are referred to as online branch and target branch), they have sub-branches $A_1, A_2$ and $B_1, B_2$. 
By introducing sub-branches, we can fuse the information on sub-branches, e.g. $A_2$ and $B_2$, before the loss end, and share this fused feature for $A_1$ and $B_1$ for further operations. In this case, the fused feature should be useful for $A_1$ and $B_1$ in generating richer latent representations. 
In the point cloud deep learning domain, local-to-global cross-attention for feature fusion has been proven to be an effective operation \cite{Engel2021PointT, pointformer, patchformer} and is ideal in our case. Therefore, we use sub-branches $A_1$ and $B_1$ for global feature learning, and sub-branches $A_2$ and $B_2$ for local feature learning.
In practice, since the patch features are from different augmentations, for one certain branch, using the local features from the other branch as the key and value input directly is not ideal for the subsequent step of local-to-global cross attention. We hence further propose an additional aligner module to align the local features from different augmentations.

As the basics of the framework have been established, the only question that remains to be answered now is: For the vanilla mode that only has two main branches, it is quite clear that the encoder on one branch updates with the gradient backpropagation normally, while the encoder on the other branch (gradient stopped) updates with momentum update. But in our framework, there are four encoders on four sub-branches, which ones should be updated normally, and which ones should be momentum updated? Our proposal and also the best practice is: follow the gradient. 
The proposed framework is illustrated in Figure \ref{fig:framework}, in which green arrows indicate the gradient flow, with red crosses indicating that gradients are stopped on these paths. In this case, we update the encoders on sub-branches $A_1, A_2$, and $B_1$ normally with backpropagation (which means they share a same encoder), while updating the encoder on sub-branch $B_2$ with momentum update.

\subsection{Overall Framework}
\label{sec:architecture}
Given a point cloud $\mathcal{P}$, the goal of PoCCA is to pre-train a powerful encoder $f_{\theta}$ that can encode it into a good latent representation that can be used for downstream tasks. There are four branches in our network architecture: online global branch, online patch branch, target global branch, and target patch branch, corresponding to the four branches in Figure \ref{fig:framework} from top to bottom. The online branch consists of five stages: (i) two sampling sub-branches to obtain a global sample with sampling function $\pi_{\text{fps}}$ and multiple local samples with sampling function $\pi_{\text{patch}}$; (ii) an encoder $f_{\theta_1}$ for shape global/patch encoding; (iii) an aligner $l_{\eta}$ for local feature aligning; (iv) a cross-attention module $g_{\phi_1}$ for merging local and global features; and (v) a predictor $h_{\psi}$. The target branch uses the same structure as the online branch but without the predictor. Moreover, while the target patch branch uses a parameter-shared encoder $f_{\theta_1}$ with the online branch, the target global branch uses another encoder $f_{\theta_2}$ whose parameters are momentum updated with the online parameters in $f_{\theta_1}$. The same goes for the target attention module $g_{\phi_2}$, whose parameters are momentum updated with the online parameters in $g_{\phi_1}$.

\subsection{PoCCA Step by Step}\label{sec:methods}

\textbf{Augmentation.} 
Many successful self-supervised learning approaches cast the prediction problem directly into representation space: the representation of one augmented view of an image should be similar to the representation of another augmented view of the same image  \cite{richemond2020byol, chen2021exploring, Chen2021CrossViTCM}. However, compared to 2D image benchmarks that have millions of training samples, 3D datasets are typically much smaller in size, and often have fewer labels and less diversity. Therefore, for 3D vision, data augmentation is a very important step to avoid overfitting and improve the generalization ability of the network. 
Same as many previous methods \cite{pointnet, pointnet++}, we augment input point clouds with random rotation, scaling, translation, and jittering.

\begin{figure}[t]
    \centering
    \includegraphics[width=\linewidth,trim=2 2 2 2,clip]{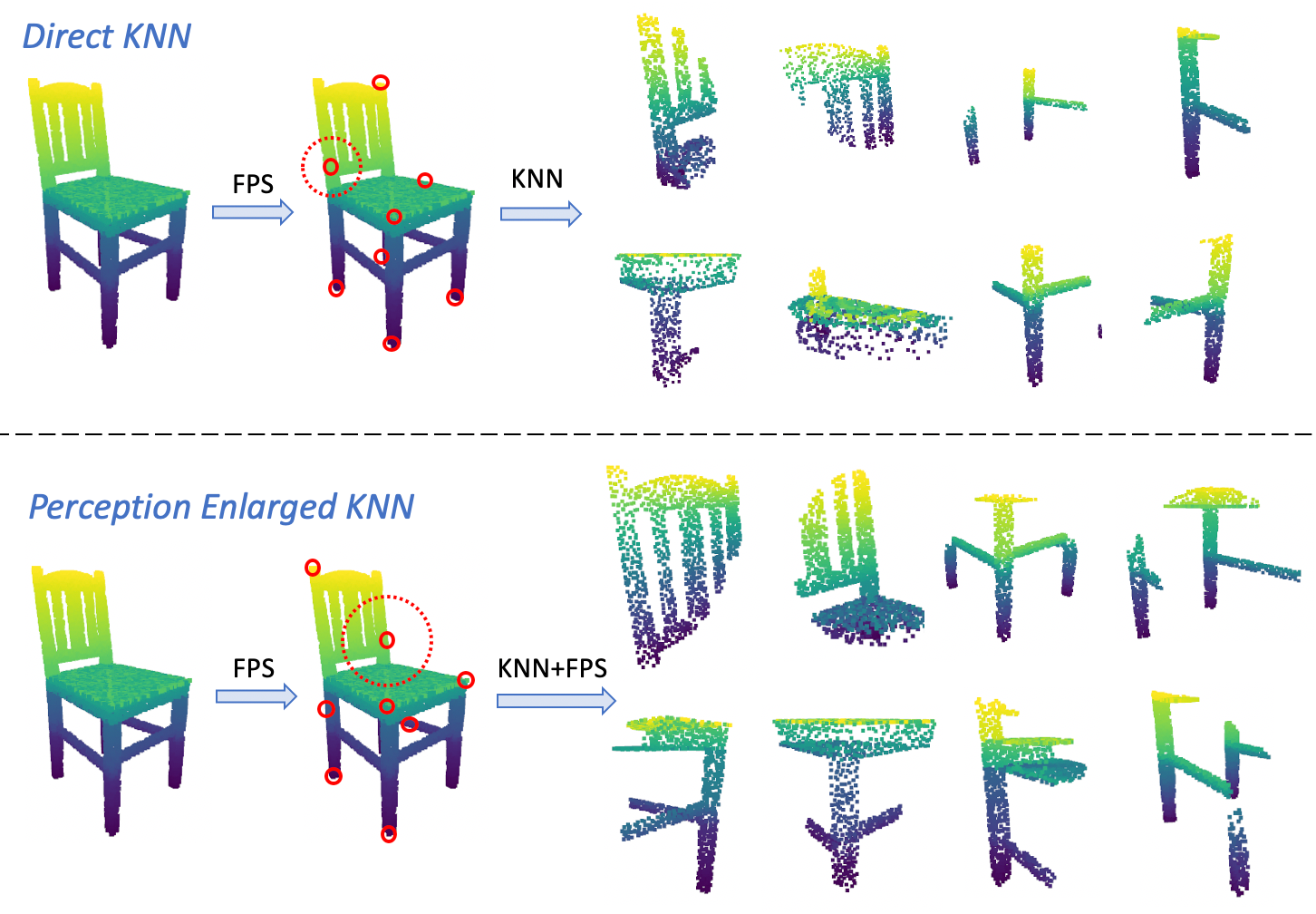}
    \caption{Perception enlarged KNN-based patch sampling. After kernel points are selected, direct-KNN gets patches with $K$neighbors for each kernel point directly, while perception-enlarged-KNN gets patches with $2^{\alpha} K$ neighbors first and then samples them to $K$ points with FPS. $\alpha$ is the scale factor.}
    \label{fig:KNNpatch}
\end{figure}

\textbf{Sampling methods.} We sample the raw point cloud to capture its global and local features, respectively. Farthest point sampling (FPS) \cite{Eldar1994TheFP} is used for global sampling.
For local sampling, given a desired number of patches $N_p$, such many kernel points are first selected with FPS on the original point cloud. Then for each kernel point, $K$ nearest neighbors are gathered to form a patch of $K$ points. In our experiments, we adopt a perception-enlarged multi-scale KNN sampling strategy. A demo is given in Figure \ref{fig:KNNpatch} for output patch comparison of direct KNN and perception-enlarged KNN. The multi-scale idea has also been used in many other advanced frameworks in other fields, e.g. GCN \cite{GCN} and Stratified Transformer \cite{Lai2022StratifiedTF}.
Denoting $\alpha$ as the scale factor, after $N_p$ kernel points are first selected, $2^{\alpha} K$ points are gathered with KNN, and subsequently downsampled to $K$ points with FPS. For scale 0 ($\alpha=0$), it falls into the normal direct KNN method. We use multi-scale $\alpha=0, 1, 2$ as the default setting.

\textbf{Encoder.} For a fair comparison with existing self-supervised methods, we use PointNet \cite{pointnet}, DGCNN \cite{Wang2019DynamicGC} and Transformer as the backbone for extracting point cloud features.  
To enable contrastive learning, the encoders for the online global branch, the online patch branch, and the target patch branch share the same weights, while the encoder for the target global branch is updated via the momentum update with the weights from other branches. Specifically, we parameterize the online branches with $\theta_1$ and the target branch with $\theta_2$. The target branch is used to train the online branches, and its parameters $\theta_2$ are an exponential moving average of the online parameters $\theta_1$.
\begin{equation}
\theta_2 \leftarrow \tau\theta_2 + (1 - \tau) \theta_1\label{eq:encoder_update}
\end{equation}
where $\tau \in (0,1)$ is the decay rate of moving average.

\begin{figure}[t]
    \centering
    \includegraphics[width=0.9\linewidth,trim=2 2 2 2,clip]{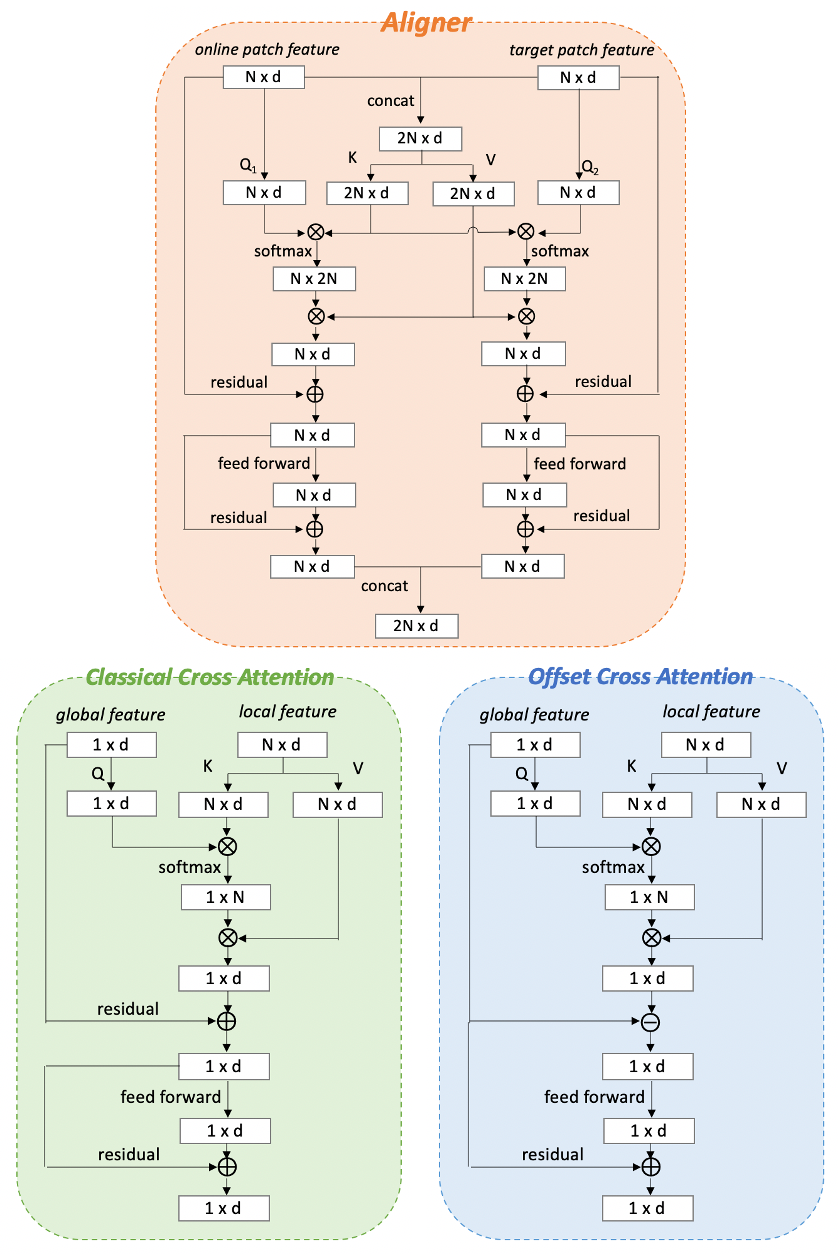}
    \caption{Detailed architecture of the aligner module and two local-to-global cross-attention variants.}
    \label{fig:attentions}
\end{figure}

\textbf{Aligner.} The aligner module is an important component of PoCCA. It enables information exchange between the online branch and the target branch. 
Since the patch features are from different augmentations, for one certain branch, using the patch features from the other branch as the key and value input directly is not ideal for the subsequent step of local-to-global cross attention. We propose to use an aligner module to align the patch features from different augmentations.
The aligner module is basically a mixture of self-attention and cross-attention by using the local patch features from two branches as the queries separately, while using them both as the key and value dictionary. Its detailed architecture is given in Figure \ref{fig:attentions}.

\textbf{Cross-Attention.} 
Cross-attention takes two separate embedding sequences of the same dimension and fuses them asymmetrically.
For local-global feature fusion, the global feature serves as the query input, while the local patch features serve as the key and value input. As with updating the encoder, we use the momentum update method to update our cross-attention. Specifically, we denote the parameters for the online cross-attention module as $\phi_1$, for the target cross-attention module as $\phi_2$, and $\tau \in (0,1)$ keeps up with the former $\tau$.
\begin{equation}
\phi_2 \leftarrow \tau\phi_2 + (1 - \tau) \phi_1\label{eq:corss_attention_update}  
\end{equation}

\textbf{Predictor.} In particular, inspired by BYOL \cite{richemond2020byol}, PoCCA uses symmetric network branches that interact and learn from each other. Since the training pair comes from the same original point cloud, it is possible for the encoder to produce the same representation for all augmented samples, which means that the network falls into a collapsed solution. In order to prevent collapse, and based on the experience of other methods \cite{richemond2020byol, chen2021exploring}, we append a predictor to all online branches, which has been shown to be effective. Our predictor is a simple MLP-based network consisting of two linear layers and a batch norm layer.  

\subsection{Loss}
\label{sec:loss}
As described in Section \ref{sec:architecture}, for a given point cloud $\mathcal{P}$, PoCCA first augments $\mathcal{P}$ to get two augmentations $m_1$ and $m_2$ by using two different augmentations. Then we process $m_1$ and $m_2$ with sampling function $\pi_{\text{fps}}$ to get two global samples $\sigma_1$, $\sigma_2$, and with sampling function $\pi_{\text{patch}}$ to get a set of local samples $\mathcal{A}=\{a_1^1, a_1^2, \dots, a_1^{N_p}, a_2^1, a_2^2, \dots, a_2^{N_p}\}$, where each $a$ denotes one certain patch. After that, $\sigma_1$ and $\mathcal{A}$ are sent to the online branch encoder $f_{\theta_1}$, while $\sigma_2$ is sent to the target branch encoder $f_{\theta_2}$. 
After being processed with the aligner module $l_{\eta}$, the outputs are subsequently sent to the respective cross-attention module $g_{\phi_1}$ and $g_{\phi_2}$.
Denote the output of the online branch as $z_{\theta_1}$ and the output of the target branch as $z_{\theta_2}$. $h_\psi$ denotes the predictor for online branches. We minimize the similarity loss between $h_\psi(z_{\theta_1})$ and $z_{\theta_2}$, which is defined by their mean square error:
\begin{equation}
\begin{split}
\mathcal{L}_{\sigma_1, \sigma_2} & \triangleq \left \| \frac{h_\psi (z_{\theta_1})}{\left\| h_\psi(z_{\theta_1})\right\|_2} -  \frac{z_{\theta_2}}{\left\|z_{\theta_2}\right\|_2}\right\|^2_2 \\
& = 2 - 2\cdot\frac{\langle h_\psi(z_{\theta_1}),  z_{\theta_2} \rangle }{\big\|h_\psi(z_{\theta_1})\big\|_2\cdot
\big\|z_{\theta_2}\big\|_2}
\end{split}
\label{eq:cosine-loss}
\end{equation}

During each training step, we swap the 2 shape augmentations for online and target branches to compute the symmetry $\mathcal{L}_{\sigma_2, \sigma_1}$. The total loss is given as:
\begin{equation}
\mathcal{L}_{total} = \mathcal{L}_{\sigma_1, \sigma_2} + \mathcal{L}_{\sigma_2, \sigma_1}
\label{eq:finalLoss}
\end{equation}

\section{Experiments}

\subsection{Pre-training}\label{sec:pretrain}
\textbf{Dataset.} We pre-train our model with the ShapeNet \cite{shapenet} dataset, which contains 57,448 synthetic models from 55 categories. The point clouds are preprocessed following the work of Yang et al. \cite{foldingnet} and each contains 2048 points. After inputting into the network, each 3D shape is first augmented with different augmentation operations.

\textbf{Implementation Details.} For each shape augmentation, we sample a global point cloud of 1024 points with FPS and local patches of 256 points (\ie, $K=256$) of multi-scales (8 patches per scale) with perception-enlarged multi-scale KNN.
We use the Adam optimizer with parameter weight decay $1\times 10^{-4}$. The initial learning rate is set as $1\times 10^{-4}$. A cosine annealing schedule is used for the learning rate and the model is trained end-to-end for $100$ epochs with a batch size of $16$. 
We set the exponential moving average parameter $\tau = 0.99$. After pre-training, the target branch encoder, alinger $l_{\eta}$, cross-attention $g_{\phi}$, and predictor $h_{\psi}$ are all discarded. All downstream tasks are performed on the pre-trained encoder $f_{\theta}$.

\begin{table}[t]
\centering
\resizebox{1\linewidth}{!}{
\begin{tabular}{l l c c c c}
\toprule
\multirow{2}{*}{Category} &
\multirow{2}{*}{Method} &
\multirow{2}{*}{Backbone} &
\multirow{2}{*}{\begin{tabular}[c]{@{}c@{}}Extra \\ Training Data\end{tabular}} &
\multicolumn{2}{c}{Overall Accuracy(\%)}
\\
\cmidrule{5-6}
  & & & & ModelNet40 & ScanObjectNN \\
\midrule
\multirow{12}{*}{\makecell[c]{\emph{Self-supervised} \\ \emph{(Reconstruction} \\ \emph{-based)}}} 
 & SO-Net  \cite{sonet} & SO-Net-Encoder & $\times$ & 87.3 & - \\
 & FoldingNet  \cite{foldingnet} & FoldingNet-Encoder & $\times$ & 88.4 & - \\
 & MRTNet  \cite{mrtnet} & MRT-Encoder & $\times$ & 86.4 & - \\
 & 3D-PointCapsNet  \cite{3dcaps} & 3D Capsule-Encoder & $\times$ & 88.9  & - \\
 & VIP-GAN  \cite{vipgan} & EncoderRNN & $\times$ & 90.2 & - \\
 & OcCo \cite{occo} & PointNet & $\times$ & 88.7 & 69.5 \\
 & OcCo \cite{occo} & DGCNN & $\times$ & 89.2 & 78.3\\
 & Point-BERT \cite{PointBERTP} & Transformer & $\times$ & 87.4 & - \\
 & Point-MAE \cite{PointMAE} & Transformer & $\times$ & 91.0 & 77.7\\
 & Point-M2AE \cite{PointM2AE} & H. Transformer & $\checkmark$ & 92.9 & 84.1 \\
 & I2P-MAE \cite{I2P-MAE} & H. Transformer & $\checkmark$ & \textbf{93.4} & \textbf{87.1} \\
 & ReCon \cite{ReCon} & Transformer & $\checkmark$ & \textbf{93.4} & - \\
\midrule
\multirow{16}{*}{\makecell[c]{\emph{Self-supervised} \\ \emph{(Pretext tasks /} \\ \emph{Contrastive}\\ \emph{-based)} \\
}} %\emph{Encoder frozen}
  &Jigsaw  \cite{jigsaw} & PointNet & $\times$ & 87.3 & 55.2\\
 & STRL  \cite{Huang2021SpatiotemporalSR} & PointNet & $\times$ & 88.3 & 74.2\\
 & Rotation  \cite{rotation} & PointNet & $\times$ & 88.6 & - \\
 & CrossPoint  \cite{Afham2022CrossPointSC} & PointNet & $\checkmark$ & 89.1 & \textbf{75.6}\\
 & SelfCorrection \cite{self_correction} & PointNet & $\checkmark$ & \textbf{89.9} & - \\
 & PoCCA (Ours) & PointNet & $\times$ & 89.4 & \textbf{75.6}\\
\cmidrule{2-6}
 & ClusterNet \cite{cluster} & DGCNN & $\times$ & 86.8 & - \\
 & Multi-Task  \cite{multitask} & DGCNN & $\times$ & 89.1 & - \\
 & Self-Contrast  \cite{acmmm} & DGCNN & $\times$ & 89.6 & - \\
 & HSN \cite{hns} & DGCNN & $\times$ & 89.6 & - \\
 & Jigsaw  \cite{jigsaw} & DGCNN & $\times$ & 90.6 & 59.5\\
 & STRL  \cite{Huang2021SpatiotemporalSR} & DGCNN & $\times$ &  90.9 & 77.9\\
 & Rotation  \cite{rotation} & DGCNN & $\times$ & 90.8 & - \\
 & CrossPoint  \cite{Afham2022CrossPointSC} & DGCNN & $\checkmark$ & 91.2 & 81.7\\
 & PoCCA (Ours) & DGCNN & $\times$ & \textbf{91.4} & \textbf{82.2}\\
 \cmidrule{2-6}
 & PoCCA (Ours) & Transformer & $\times$ &\textbf{92.1} & \textbf{83.6}\\
\bottomrule
\end{tabular}}
\caption{Comparison of linear classification results with previous self-supervised methods on ModelNet40 and ScanObjectNN. A linear classifier is fit onto the shape global representation learned with the pre-trained model (the encoder is frozen). The overall accuracy for the classification task is reported. 
"H. Transformer" stands for Hierarchical Transformer. 
PoCCA achieves state-of-the-art results among the methods that do not use extra training data in both backbones.}
\label{table:cls_svm}
\end{table}

\subsection{Justification of Performed Comparison}
\label{sec:justification}
Although in various fields nowadays, multi-modal approaches often outperform their single-modal counterparts in terms of performance, and this study primarily focuses on research involving single-modal data, we have still included the results of some recent multi-modal methods for comparison. This is because they remain important relevant literature in the field of unsupervised learning. 
On the other hand, please note that for self-supervised learning frameworks, the key is the framework itself, other than the backbone. That is why MoCo, SimCLR, BYLO, etc. all used ResNet-50 as the backbone for image self-supervised contrastive learning. Using a more advanced backbone would surely improve performance, but they did not for a fair comparison. But for those reconstruction-based self-supervised learning methods, various customized backbones are used since their frameworks include additional generators, especially the recent MAE-based ones.
Before these MAE-based methods, in the point cloud contrastive learning sub-domain, most papers use PointNet and DGCNN as the backbone for a fair framework comparison. In our experiments, we mainly compare our results with those that use PointNet and DGCNN as the backbone, but meanwhile, the results on a simple Transformer backbone is also reported.

\subsection{Downstream Tasks}
\label{sec:downstream}
\textbf{Linear SVM Classification.}
For the classification task, we adopt the protocols of previous work \cite{Huang2021SpatiotemporalSR} to evaluate the transferability of PoCCA on the ModelNet40 \cite{modelNet} and ScanObjectNN \cite{scanobjectnn} benchmarks. A linear Support Vector Machine (SVM) \cite{svm} is used to classify 3D shapes by applying it to the encoded global feature representations. We freeze the pre-trained encoder and fit a simple linear SVM classifier on the train split of ModelNet40 and ScanObjectNN, respectively. 
Experimental results are presented in Table \ref{table:cls_svm}. Note that reconstruction-based methods typically design specific encoder-decoder architectures and do not use common encoders (e.g. PointNet, DGCNN) as the backbone. They also typically incorporate an additional reconstruction loss.
As shown in Table \ref{table:cls_svm}, in the category of pretext tasks and contrastive-based self-supervised learning method, PoCCA outperforms other state-of-the-art unsupervised methods with both backbones. It even outperforms the reconstruction-based methods that do not use extra training data.

\begin{table}[t]
\centering
\resizebox{1\linewidth}{!}{
\begin{tabular}{c l c c c c}
\toprule
\multirow{2}{*}{Category} &
\multirow{2}{*}{Method} & 
\multirow{2}{*}{Backbone} & 
\multirow{2}{*}{\begin{tabular}[c]{@{}c@{}}Extra \\ Training Data\end{tabular}} &
\multicolumn{2}{c}{Overall Accuracy(\%)}
\\
\cmidrule{5-6}
 & & & & ModelNet40 & ScanObjectNN \\
\midrule
\multirow{5}{*}{\emph{Supervised}} 
 & PointNet~\cite{pointnet} & - & $\times$ & 89.2 & 68.2\\ 
 & PointNet++~\cite{pointnet++} & - & $\times$ & 90.7 & 77.9\\ 
 & PointCNN~\cite{pointcnn} & - & $\times$ & 92.2 & 78.5\\ 
 & DGCNN~\cite{Wang2019DynamicGC} & - & $\times$ & 92.9 & 78.1\\
 & PCT~\cite{Guo2021PCTPC} & - & $\times$ & 93.2 & - \\
 \midrule
\multirow{7}{*}{\makecell[c]{\emph{Self-supervised} \\ \emph{(Reconstruction} \\ \emph{-based)} \\
}} 
 & OcCo  \cite{occo} & PointNet & $\times$ & 90.1 & 80.0\\
 & OcCo~\cite{occo}  & DGCNN & $\times$ & 93.0 & 83.9\\ 
 & Point-BERT \cite{PointBERTP} & Transformer & $\times$ & 92.7 & 83.1 \\
 & Point-MAE \cite{PointMAE} & Transformer & $\times$ & 93.2 & 85.2 \\
 & Point-M2AE \cite{PointM2AE} & H. Transformer & $\checkmark$ & 93.4 & - \\
 & I2P-MAE \cite{I2P-MAE} & H. Transformer & $\checkmark$ & \textbf{93.7} & 90.1 \\
 & ReCon \cite{ReCon} & Transformer & $\checkmark$ & \textbf{94.1} & \textbf{90.6} \\
 \midrule
\multirow{11}{*}{\makecell[c]{\emph{Self-supervised} \\ \emph{(Pretext tasks /} \\ \emph{Contrastive} \\ \emph{-based)} \\
}} %\emph{Encoder frozen}
 & Jigsaw  \cite{jigsaw} & PointNet & $\times$ & 89.6 & 76.5\\
 & Info3D ~\cite{info3d} & PointNet & $\times$ & 90.2 & - \\
 & SelfCorrection \cite{self_correction} & PointNet & $\checkmark$ & 90.0 & - \\
 & ParAE~\cite{ParAE}  & PointNet & $\checkmark$ & \textbf{90.5} & - \\ 
 & PoCCA (Ours) & PointNet & $\times$ & 90.2 & \textbf{80.3} \\
\cmidrule{2-6}
 & Jigsaw~\cite{jigsaw} & DGCNN & $\times$ & 92.4 & 82.7 \\
 & ParAE~\cite{ParAE}  & DGCNN & $\checkmark$ & 92.9 & - \\ 
 & Info3D~\cite{info3d} & DGCNN & $\times$ & 93.0 & - \\ 
 & STRL~\cite{Huang2021SpatiotemporalSR}  & DGCNN & $\times$ &  93.1 & - \\ 
 & PoCCA (ours) & DGCNN & $\times$ & \textbf{93.2} &  \textbf{84.1} \\
 \cmidrule{2-6}
 & PoCCA (ours) & Transformer & $\times$ & \textbf{93.3} &  \textbf{84.8} \\
\bottomrule
\end{tabular}}
\caption{Results of shape classification task network fine-tuned on ModelNet40 and ScanObjectNN. The self-supervised pre-trained backbone encoders serve as the initial weights for supervised downstream tasks.}
\label{table:cls_finetune}
\end{table}

\begin{figure}[t]
    \centering
    \includegraphics[width=\linewidth,trim=2 2 2 2,clip]{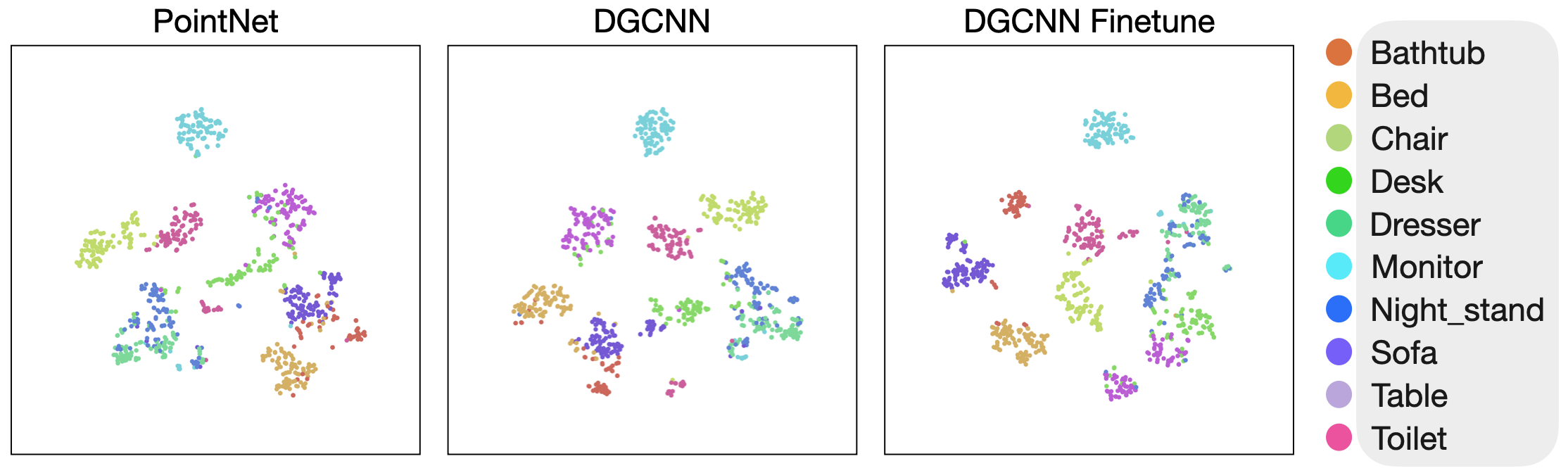}
    \caption{t-SNE visualization of features on the test split of ModelNet10 dataset, with PointNet and DGCNN as the backbone in a self-supervised manner (left and middle), the right one illustrates the latent features learned with DGCNN after fine-tuning.}
    \label{fig:tSNE}
\end{figure}

\textbf{Fine-tuned Classification.}
The results of transfer learning on the ModelNet40 classification task are reported in Table \ref{table:cls_finetune}, \ie, the decoder is first initialized with pre-trained weights, then the whole task network is fine-tuned in a supervised manner on the training set. From it, we observe that our method outperforms most other SOTA contrastive-based methods. 
The t-SNE plots are given in Figure \ref{fig:tSNE} for better visualization of the learned latent representations.
%-------- finetuning on partial samples -----------
Additional results of fine-tuning on partial training samples are also provided in Table \ref{tab:dgcnncls_few} in comparison with STRL~\cite{Huang2021SpatiotemporalSR}.

\begin{table}[t]
    \centering
        \scalebox{0.6}{
            \begin{tabular}{lccccc}
                \toprule
                Method &  1\% & 5\%  & 10\% & 20\% & 100\% \\
                \midrule
                DGCNN & 58.4\% & 80.7\% & 85.2\% & 88.1\% & 92.9\%\\ \midrule
                STRL + DGCNN & 60.5\% & 82.7\% & 86.5\% & 89.7\% & 93.1\%\\
                PoCCA + DGCNN & \textbf{65.7\%} & \textbf{85.8}\% & \textbf{89.2}\% & \textbf{91.3}\% & \textbf{93.2}\% \\
                \bottomrule 
            \end{tabular}%
        }%
        \caption{Shape classification fine-tuned on ModelNet40 with only partial training samples.}
        \label{tab:dgcnncls_few}
\end{table}

\textbf{Few-shot Classification.}
We additionally give the results on the few-shot object classification metric. Few-shot learning (FSL) \cite{fsl} is a machine learning methodology where models are trained on small datasets, where each category only provides a few instances. 
We test our models on a standard few-shot task, namely X-way Y-shot learning, where the model is evaluated on X classes, and each class contains Y samples. Like standard 3D object classification. We again use ModelNet40 \cite{modelNet} and ScanObjectNN \cite{scanobjectnn} datasets to carry out FSL experiments. We take 10 random few-shot tasks and report the mean and standard deviation of their results. As presented in Table \ref{tab:fsl}, our FSL results on ModelNet40 show that PoCCA outperforms most prior works for both PointNet and DGCNN backbones. The FSL results for the ScanObjectNN dataset are presented in Table \ref{tab:scan_fsl}. By using PoCCA, the accuracy is also increased significantly in most settings for both feature extractors.

\begin{table}[t]
\centering
\setlength{\tabcolsep}{3pt}
\resizebox{0.9\linewidth}{!}{
\begin{tabular}{l|cccc}
\toprule \multicolumn{1}{c|}{\multirow{2}{*}{Method}}  & 
\multicolumn{2}{c}{5-way} & \multicolumn{2}{c}{10-way} \\\cmidrule{2-5}
& 10-shot & 20-shot & 10-shot & 20-shot \\ \midrule \midrule
3D-GAN  \cite{3dgan} & 55.8$\pm$3.4 & 65.8$\pm$3.1 & 40.3$\pm$2.1 & 48.4$\pm$1.8 \\
FoldingNet  \cite{foldingnet}  & 33.4$\pm$4.1 & 35.8$\pm$5.8 & 18.6$\pm$1.8 & 15.4$\pm$2.2 \\
Latent-GAN  \cite{latentgan}  & 41.6$\pm$5.3 & 46.2$\pm$6.2 & 32.9$\pm$2.9 & 25.5$\pm$3.2 \\
3D-PointCapsNet  \cite{3dcaps} & 42.3$\pm$5.5 & 53.0$\pm$5.9 & 38.0$\pm$4.5 & 27.2$\pm$4.7 \\
PointNet++  \cite{pointnet++}  & 38.5$\pm$4.4 & 42.4$\pm$4.5 & 23.1$\pm$2.2 & 18.8$\pm$1.7 \\
PointCNN  \cite{pointcnn}& 65.4$\pm$2.8 & 68.6$\pm$2.2 & 46.6$\pm$1.5 & 50.0$\pm$2.3 \\
RSCNN  \cite{rscnn} & {65.4$\pm$8.9} & {68.6$\pm$7.0} & {46.6$\pm$4.8} & {50.0$\pm$7.2} \\ \midrule
PointNet + Rand & 52.0$\pm$3.8 & 57.8$\pm$4.9 & 46.6$\pm$4.3 & 35.2$\pm$4.8 \\
PointNet + Jigsaw  \cite{jigsaw} & 66.5$\pm$2.5 & 69.2$\pm$2.4 & 56.9$\pm$2.5 & 66.5$\pm$1.4\\
PointNet + cTree  \cite{ctree}& 63.2$\pm$3.4 & 68.9$\pm$3.0 & 49.2$\pm$1.9 & 50.1$\pm$1.6 \\
PointNet + OcCo  \cite{occo}& 89.7$\pm$1.9 & 92.4$\pm$1.6 & 83.9$\pm$1.8 & 89.7$\pm$1.5 \\
PointNet + CrossPoint  \cite{Afham2022CrossPointSC}& 90.9$\pm$4.8 & 93.5$\pm$4.4 & 84.6$\pm$4.7 & 90.2$\pm$2.2 \\
PointNet + PoCCA (Ours) & \textbf{91.7$\pm$3.1} &\textbf{94.2$\pm$3.5} & \textbf{87.3$\pm$2.9} & \textbf{90.9$\pm$4.1} \\ \midrule
DGCNN + Rand & 31.6$\pm$2.8 & 40.8$\pm$4.6 & 19.9$\pm$2.1 & 16.9$\pm$1.5\\
DGCNN + Jigsaw  \cite{jigsaw}& 34.3$\pm$1.3 & 42.2$\pm$3.5 & 26.0$\pm$2.4 & 29.9$\pm$2.6\\
DGCNN + cTree  \cite{ctree} & 60.0$\pm$2.8 & 65.7$\pm$2.6 & 48.5$\pm$1.8 & 53.0$\pm$1.3\\
DGCNN + OcCo  \cite{occo} & 90.6$\pm$2.8 & 92.5$\pm$1.9 & 82.9$\pm$1.3 & 86.5$\pm$2.2 \\
DGCNN + CrossPoint  \cite{Afham2022CrossPointSC} & 92.5$\pm$3.0 &  \textbf{94.9$\pm$2.1} & 83.6$\pm$5.3 & 87.9$\pm$4.2 \\
DGCNN + PoCCA (Ours) & \textbf{93.5$\pm$3.7} & 92.1$\pm$3.6 & \textbf{88.1$\pm$5.3} & \textbf{90.9$\pm$4.0}\\
\midrule \bottomrule
\end{tabular}}
\caption{Few-shot object classification results on ModelNet40. We report mean and standard error over 10 runs. Our proposed PoCCA improves the few-shot accuracy in most of the reported settings. Table is extended from  \cite{Afham2022CrossPointSC}.}
\label{tab:fsl}
\end{table}

\begin{table}[t]
\centering
\setlength{\tabcolsep}{3pt}
\resizebox{0.9\linewidth}{!}{
\begin{tabular}{l|cccc}
\toprule \multicolumn{1}{c|}{\multirow{2}{*}{Method}}  & 
\multicolumn{2}{c}{5-way} & \multicolumn{2}{c}{10-way} \\ \cmidrule{2-5}
& 10-shot & 20-shot & 10-shot & 20-shot \\ \midrule \midrule
PointNet + Rand & 57.6$\pm$2.5 & 61.4$\pm$2.4 & 41.3$\pm$1.3 & 43.8$\pm$1.9 \\
PointNet + Jigsaw  \cite{jigsaw} & 58.6$\pm$1.9 & 67.6$\pm$2.1 & 53.6$\pm$1.7 & 48.1$\pm$1.9\\
PointNet + cTree  \cite{ctree}& 59.6$\pm$2.3 & 61.4$\pm$1.4 & 53.0$\pm$1.9 & 50.9$\pm$2.1 \\
PointNet + OcCo  \cite{occo}& 70.4$\pm$3.3 & 72.2$\pm$3.0 & 54.8$\pm$1.3 & 61.8$\pm$1.2 \\
PointNet + CrossPoint  \cite{Afham2022CrossPointSC}& 68.2$\pm$3.3 & 73.2$\pm$2.9 & 58.7$\pm$1.8 & 64.6$\pm$1.2 \\
PointNet + PoCCA (Ours) & \textbf{70.5$\pm$1.8} & \textbf{74.8$\pm$3.2} & \textbf{60.3$\pm$2.3} & \textbf{65.2$\pm$1.7} \\ \midrule
DGCNN + Rand & 62.0$\pm$5.6 & 67.8$\pm$5.1 & 37.8$\pm$4.3 & 41.8$\pm$2.4\\
DGCNN + Jigsaw  \cite{jigsaw}& 65.2$\pm$3.8 & 72.2$\pm$2.7 & 45.6$\pm$3.1 & 48.2$\pm$2.8\\
DGCNN + cTree  \cite{ctree} & 68.4$\pm$3.4 & 71.6$\pm$2.9 & 42.4$\pm$2.7 & 43.0$\pm$3.0\\
DGCNN + OcCo  \cite{occo} & 72.4$\pm$1.4 & 77.2$\pm$1.4 & 57.0$\pm$1.3 & 61.6$\pm$1.2\\
DGCNN + CrossPoint  \cite{Afham2022CrossPointSC} & 74.8$\pm$1.5 & 79.0$\pm$1.2 & 62.9$\pm$1.7 & 73.9$\pm$2.2\\
DGCNN + PoCCA (Ours) & \textbf{79.9$\pm$4.7} &\textbf{83.5$\pm$4.2} & \textbf{66.0$\pm$3.2} & \textbf{75.1$\pm$2.7}\\
\midrule 
\bottomrule
\end{tabular}}
\caption{Few-shot object classification results on ScanObjectNN. We report mean and standard error over 10 runs. Our proposed PoCCA improves the few-shot accuracy in all the reported settings. Table is extended from \cite{Afham2022CrossPointSC}.}
\label{tab:scan_fsl}
\end{table}

\textbf{Part Segmentation.} 
The pre-trained model is also used for the 3D part segmentation task on the ShapeNetPart \cite{shapenet_part} dataset. This dataset contains a total of 16881 3D objects from 16 different categories with 50 annotated semantic parts. We first pre-train our PoCCA framework with the DGCNN or Transformer backbone on the ShapeNet \cite{shapenet} dataset. The model is then fine-tuned on the training set of ShapeNetPart. 
Both category mIoU and instance mIoU are computed and presented in Table \ref{tab:partseg}. 
It shows that the backbone pre-trained via PoCCA leads to better part segmentation performance than other self-supervised methods, as well as the randomly initialized DGCNN baseline. 
Regarding the transformer backbone, PoCCA does not achieve comparable results to methods that use extra training data, yet still outperforms the ones that do not.
Overall, PoCCA is a good choice for weight initialization for feature extractors, as proved by the results.

\begin{table}[t]
\centering
\resizebox{1\linewidth}{!}{
\begin{tabular}{c l c c c c}
\toprule
Category & Method & Backbone & \begin{tabular}[c]{@{}c@{}}Extra \\ Training Data\end{tabular} & Cat. mIoU & Ins. mIoU \\
\midrule
\multirow{5}{*}{\emph{Supervised}} 
 & PointNet~\cite{pointnet} & - & $\times$ & 80.4 & 83.7\\ 
 & PointNet++~\cite{pointnet++} & - & $\times$ & 81.9 & 85.1\\ 
 & DGCNN~\cite{Wang2019DynamicGC} & - & $\times$ & 82.3 & 85.1\\
 & RSCNN~\cite{rscnn} & - & $\times$ & 84.0 & 86.2\\
 & PCT~\cite{Guo2021PCTPC} & - & $\times$ & 83.1 & 86.4 \\
\midrule
\multirow{4}{*}{\makecell[c]{\emph{Self-supervised} \\ \emph{(Encoder frozen)} \\
}}
 & CloudContext \cite{cloudcontext} & DGCNN & $\times$ & - & 81.5 \\
 & HNS \cite{hns} & DGCNN & $\times$ & 79.9 & 82.3 \\
 & CMCV \cite{cmcv} & DGCNN & $\times$ & 74.7 & 80.8 \\
 & PoCCA (Ours) & DGCNN & $\times$ & \textbf{80.8} & \textbf{83.7} \\
\midrule
\multirow{11}{*}{\makecell[c]{\emph{Self-supervised} \\ \emph{(Encoder} \\ \emph{fine-tuned)} \\
}} 
 & Jigsaw \cite{jigsaw} & DGCNN & $\times$ & 83.1 & 85.3 \\
 & CMCV \cite{cmcv} & DGCNN & $\times$ & 79.1 & 83.7 \\
 & OcCo \cite{occo} & DGCNN & $\times$ & 84.4 & - \\
 & CrossPoint \cite{Afham2022CrossPointSC} & DGCNN & $\checkmark$ & - & 85.5 \\
 & PoCCA (Ours)  & DGCNN & $\times$ & \textbf{84.5} & \textbf{85.8} \\
\cmidrule{2-6}
 & Point-BERT \cite{PointBERTP} & Transformer & $\times$ & 84.1 & 85.6 \\
 & Point-MAE \cite{PointMAE} & Transformer & $\times$ & - & 86.1 \\
 & Point-M2AE \cite{PointM2AE} & H. Transformer & $\checkmark$ & 84.9 & 86.5 \\
 & I2P-MAE \cite{I2P-MAE} & H. Transformer & $\checkmark$ & \textbf{85.1} & \textbf{86.7} \\
 & ReCon \cite{ReCon} & Transformer & $\checkmark$ & 84.8 & 86.4 \\
 & PoCCA (Ours) & Transformer & $\times$ & 84.7 & 86.1 \\
\bottomrule
\end{tabular}}
\caption{Shape part segmentation results on the ShapeNetPart dataset using DGCNN or Transformer as the backbone. Both category mIoU and instance mIoU are reported. Cases of encoder frozen and encoder fine-tuned are both presented. "H. Transformer" stands for Hierarchical Transformer.}
\label{tab:partseg}
\end{table}

\subsection{Ablation Study}
\label{sec:ablation}
\textbf{PoCCA variants.} 
Apart from the PoCCA framework used above, its variants of modifying different parts are additionally investigated, including discarding local branches, not aligning local features, using different local-global feature fusion methods, and not using the predictor. 
Numerical results are given in Table\ref{table:sum_ablation}. From it, we can observe that not using local patches decreases performance significantly. Meanwhile, not merging the patch features from both branches also decreases the performance in most cases. Regarding the local-global feature fusion operation, the cross-attention module outperforms direct concatenation significantly. Direct concatenation even performs worse than not using the patches, indicating the importance of feature fusion. Apart from the classical cross-attention from Transformers, we also have tried the offset attention which was introduced in PCT \cite{Guo2021PCTPC}. Their structures are given in Figure \ref{fig:attentions}. However, while they claimed better performance in supervised learning tasks with offset attention, we observe a performance decrease in our self-supervised learning task. 
Moreover, using momentum update is helpful to the model, yet should be used wisely. We recommend following the actual backpropagated gradients.
Last but not least, same as in BYOL, when the predictor is not used, the pipeline collapses to a minimal solution and thus does not work anymore.

\textbf{Patch Sampling Methods.} 
Ablation experiments are carried out for the comparison of different patch sampling methods. All experiments are conducted with the DGCNN backbone.
From Table \ref{table:randomFPS}, we can observe that cuboid-cut and sphere-cut sampling methods achieve similar performances, which is quite reasonable since they produce similar patches. Meanwhile, the KNN-based sampling methods outperform shape-cut-based sampling methods. The additional results of using different settings of the KNN-based sampling methods show that random kernel point selection achieves lower performance, especially when the perceptual field is small. This is because patches that are not around the point cloud contours usually do not contain too much information, let alone it is hard to guarantee a good coverage of shapes with these patches. On the other hand, FPS-based kernel point selection assures better patch acquisition. When multi-scale perception KNN is used to create multi-scale patches, our method achieves the best performance.

\begin{table}[t]
\centering
\resizebox{1\linewidth}{!}{
\begin{tabular}{cccccc}
\toprule
Sub-branch & \multicolumn{1}{c}{\begin{tabular}[c]{@{}c@{}}Momentum Updated \\ Encoder Branch\end{tabular}} & \multicolumn{1}{c}{\begin{tabular}[c]{@{}c@{}}Sub-branch \\ Merge\end{tabular}} & \multicolumn{1}{c}{\begin{tabular}[c]{@{}c@{}}Local-Global \\ Merge\end{tabular}} & \multicolumn{1}{c}{Predictor} & \multicolumn{1}{c}{Accuracy} \\
\midrule
 \checkmark & Target global & Aligner & Classical CA & \checkmark & \textbf{91.4} \\
 \checkmark & Target global & Concat. & Classical CA & \checkmark & 91.0 \\
 \checkmark & Target global & - & Classical CA & \checkmark & 89.7 \\
 \checkmark & Target global & Aligner & Offset CA & \checkmark & 91.2 \\
 \checkmark & Target global & Concat. & Offset CA & \checkmark & 90.9 \\
 \checkmark & Target global & - & Offset CA & \checkmark & 89.5 \\
 \checkmark & Target global & Aligner & Concat. & \checkmark & 84.2 \\
 \checkmark & Target global & Concat. & Concat. & \checkmark & 84.3 \\
 \checkmark & Target global & - & Concat. & \checkmark & 86.1 \\
 \checkmark & None & Aligner & Classical CA & \checkmark & 90.2 \\
 \checkmark & Target both & Aligner & Classical CA & \checkmark & 85.8 \\ 
 \checkmark & Target global & Aligner & Classical CA & - &  8.3 \\
 - & Target & - & - & - & 7.7 \\
 - & Target & - & - & \checkmark & 89.6 \\
\bottomrule
\end{tabular}}
\caption{Ablation study of different modules or settings in PoCCA. 
"CA" stands for Cross-Attention.}
\label{table:sum_ablation}
\end{table}

\begin{table}[t]
\centering
\resizebox{0.65\linewidth}{!}{
\begin{tabular}{l c c} 
 \toprule 
 \multicolumn{1}{c}{\begin{tabular}[c]{@{}c@{}}Patch Sampling \\ Method\end{tabular}} & \multicolumn{1}{c}{\begin{tabular}[c]{@{}c@{}}Kernel Points \\ for KNN\end{tabular}} & \multicolumn{1}{c}{ Test Accuracy} \\
 \midrule
 Slice-cut  & - & 88.7 \\
 Cuboid-cut & - & 90.3 \\
 Sphere-cut & - & 90.5 \\
 KNN scale 0 & random & 89.2 \\
 KNN scale 1 & random & 90.1 \\
 KNN scale 2 & random & 90.5  \\
 KNN scale 0, 1, 2 & random &  90.9 \\
 KNN scale 0 & FPS & 89.6 \\
 KNN scale 1 & FPS & 90.5\\
 KNN scale 2 & FPS & 91.1 \\
 KNN scale 0, 1, 2 & FPS &  \textbf{91.4}\\
 \bottomrule
\end{tabular}}
\caption{Numerical results with different patch sampling methods. }
\label{table:randomFPS}
\end{table}

\section{Conclusion} 
In this paper, we propose an effective unsupervised framework PoCCA for point cloud representation learning. Compared to common contrastive learning frameworks, PoCCA enables information exchange between the online branch and the target branch by leveraging the local and global features of different sub-branches. We have evaluated our approach on point cloud classification and segmentation benchmarks, and the experimental results show that it achieves state-of-the-art performance between the point cloud contrastive learning methods that do not use extra training data. We have also evaluated the influence of different components of PoCCA through ablation studies. For future work, it would be interesting to investigate point-wise contrastive frameworks. New losses could be designed for better model pre-training. Moreover, better ways to exploit the attention mechanism could be explored.

\section*{Acknowledgements}
This research is funded by the Carl-Zeiss Foundation.

{
    \small
    \bibliographystyle{ieeenat_fullname}
    \bibliography{main}
}

\end{document}